# MEDAVET: Traffic Vehicle Anomaly Detection Mechanism based on spatial and temporal structures in vehicle traffic

**Ana Rosalía Huamán Reyna**  [ Universidade de São Paulo | *arhuamanr@usp.br* ]

**Alex Josué Flórez Farfán**  [ Universidade de São Paulo | *alex.josueff@usp.br* ]

**Geraldo P. Rocha Filho**  [ State University of Southwest Bahia | *geraldo.rocha@uesb.edu.br* ]

**Sandra Sampaio**  [ University of Manchester | *s.sampaio@manchester.ac.uk* ]

**Robson de Grande**  [ Brock University | *rdegrande@brocku.ca* ]

**Luis Hideo Vasconcelos Nakamura**  [ Universidade de São Paulo | *nakamura@icmc.usp.br* ]

**Rodolfo Ipolito Meneguette**  [ Universidade de São Paulo | *meneguette@icmc.usp.br* ]

✉ *Institute of Mathematical and Computer Sciences, University of São Paulo (USP)*
*Av. Trabalhador São Carlense, 400 - Centro, São Carlos - SP, 13566-590, Brazil.*

**Abstract** Currently, there are computer vision systems that help us with tasks that would be dull for humans, such as surveillance and vehicle tracking. An important part of this analysis is to identify traffic anomalies. An anomaly tells us that something unusual has happened, in this case on the highway. This paper aims to model vehicle tracking using computer vision to detect traffic anomalies on a highway. We develop the steps of detection, tracking, and analysis of traffic: the detection of vehicles from video of urban traffic, the tracking of vehicles using a bipartite graph and the Convex Hull algorithm to delimit moving areas. Finally for anomaly detection we use two data structures to detect the beginning and end of the anomaly. The first is the QuadTree that groups vehicles that are stopped for a long time on the road and the second that approaches vehicles that are occluded. Experimental results show that our method is acceptable on the Track4 test set, with an $F1$ score of 85.7% and a mean squared error of 25.432.

## 1 Introduction

Traffic safety is a topic of fundamental importance in everyday life around the world. According to data released by the World Health Organization in the Global Status Report on Road Safety 2018 [of Health, 2018], annually traffic accidents result in approximately 1.35 million deaths in worldwide. Among non-fatal victims, between 20 and 50 million people suffer from permanent sequelae and more than half (54%) of all traffic deaths and injuries involve vulnerable users, such as pedestrians, cyclists, motorcyclists, and their passengers.

These data caused intense mobilization in the global community, with the aim of developing action plans to increase street safety and, consequently, reduce the number of traffic victims. An example of an initiative in this sense is the Second Decade of Action for Road Safety [NATIONS., 2020] defined by the UN General Assembly, which aims to reduce traffic injuries and deaths by at least 50% worldwide between 2021 and 2030.

To guarantee the safety and daily operations of vehicular flow, authorities need to monitor and inspect traffic on the roads, analyzing the data generated by the installed cameras [Ferrante *et al.*, 2021]. The control they implement is related to controlling speed limits, presence of abnormal situations, detection of stopped vehicles, among others, to avoid congestion on highways [Gomides *et al.*, 2022]. Which means that it is necessary to monitor and detect and track the flow of vehicles for the normal development of society's activities [Huk and Kurowski, 2022a].

Perception systems based on camera data are strongly influenced by adverse conditions [Ge *et al.*, 2023]. Therefore, although image data has been the basis for important advances in the field of Computer Vision in recent years, it is necessary to study new perception strategies [Liu *et al.*, 2023].

Computer vision is a field of artificial intelligence research in which computers, using data from digital images, videos and other visual inputs, can obtain significant information to carry out actions or recommendations based on this data [Shabbir and Anwer, 2018]. Computer vision has demonstrated its popularity with the growing number of applications in industry, such as robotic vision, human-machine interfaces, information retrieval, medical image analysis, security systems, traffic surveillance systems, among others [Santos, 2014].

In the specific case of traffic surveillance, detecting and tracking moving vehicles is a key step for this type of computer vision system [Montanari, 2016]. Therefore, it is necessary to monitor, detect and track the flow of vehicles for the normal development of society's activities [Huk and Kurowski, 2022b].

It is necessary to control the flow of vehicles, numerous surveillance cameras have been placed on highways in order to observe and take care of the well-being of people in front of cars [Martinez *et al.*, 2010]. To deal with a large amount of vehicular traffic flow data, researchers have focused on computer vision systems to analyze and understand this data [Pawar and Attar, 2021]. Highway video data is used to obtain traffic information, such as determining the number and speed of cars or other vehicles on the highway, determining the presence of traffic anomalies, or warning of possible congestion [Djenouri *et al.*, 2022].

Consequently, the analysis and detection of traffic anomalies represent an important step in this process. An anomaly indicates that something unforeseen is happening on the highway, and therefore, it is necessary to be able to identify the anomalies to take appropriate action to the situation, and thus, for example, avoid possible accidents [Zhang *et al.*, 2021a]. Anomaly detection in road traffic is a fundamental computer vision task and plays a critical role in video structure analysis and urban traffic analysis [Zhao, 2021]. Due to the complexity of traffic conditions, doing traffic analysis remains a very challenging problem due to the complexity of the traffic scenario, the variation in the orientation of moving objects, changes in weather conditions, appearance of moving objects and non-moving objects. target in the background [Pawar and Attar, 2021].

Still in relation to vehicular traffic, many situations can occur that are not foreseen by drivers. One way to help identify these situations on the highway is by using traffic surveillance systems, and computer vision systems that can analyze and determine vehicle flow in real time. With the help of these systems, authorities can make quick and better decisions to correct the situation, as well as predict and avoid possible similar situations in the future.

Within this context, this article proposes a vehicle tracking model that uses computer vision to detect traffic anomalies on highways. The idea is to analyze vehicle traffic from urban traffic video scenes, detect vehicles, and track them, both those moving and those that are stationary, with the aim of detecting anomalies in highway traffic. For this work, an anomaly is defined as a vehicle stopped for a period longer than that established by the traffic light on a main road. Therefore, the main contributions are:

- Develop and evaluate an anomaly detection mechanism.
- Develop and evaluate a vehicle detection and tracking mechanism.
- Use bipartite graphs and time/spatial positioning to optimize the anomaly detection process and vehicle tracking.

This article is structured as follows: in section 2, some work related to the topic of this paper is analyzed and some comparisons are made. Section 3 presents the proposal developed for detecting anomalies in vehicle traffic on highways. In section 4, the results of the developed detection model are analyzed and discussed. Finally, in section 5, final considerations are made on the proposal presented.

The source code for this work is available at `https://github.com/AnaHReyna/TrackingVehicles.git`. The repository contains the tracking and anomaly detection code presented in section 3 and the setup to perform the experiments presented.

# 2 Related Work

The use of computer vision to solve the most diverse problems is quite broad. For this paper, we researched some related works that applied computer vision to solve problems related to vehicle traffic.

In [Bafghi and Shoushtarian, 2020], the objective was to present a system for tracking multiple vehicles on a highway using appearance models and visual tracking. The method consists of three steps: vehicle detection, vehicle appearance modeling and vehicle tracking. In the first step, the Mask R-CNN detection algorithm [He *et al.*, 2017] identifies the presence of vehicles in each frame of the video. In the second step, the *SIFT* [Lowe, 2004] appearance model and color histograms are created based on its visual characteristics from a reference image of each vehicle. In the third step, appearance models are used to identify and track vehicles detected in multiple subsequent images over time. To achieve this, edge detection and visual feature matching techniques are used, such as bipartite graphs. In order to obtain greater accuracy in estimating the positions of objects, a motion model was used. With these two models, appearance and motion, the edge weights are found in linear combination of these two models. The method is evaluated on the UA-DETRAC dataset and shows promising results in terms of accuracy and efficiency in detecting and tracking multiple vehicles moving on the highway.

In [Bai *et al.*, 2019], an anomaly detection system is proposed that includes three modules: Background modeling module, Perspective detection module, and Spatio-temporal matrix discrimination module. Background modeling analyzes the traffic flow to obtain the Unsupervised road segmentation results based on traffic flow analysis that eliminates interference from off-road vehicles. The detection perspective model obtains the perspective map by the first detection result, which is done by the Faster R CNN model [Fan *et al.*, 2016], and together the image is cropped into a uniform scale for different vehicles and re-detection. Finally, all anomalies are obtained by building a spatial-temporal information matrix with the detection results. Furthermore, all anomalies are combined through NMS and the re-identification model, including spatial and temporal dimensions.

In [Li *et al.*, 2020], a traffic anomaly detection method based on vehicle detection and tracking was developed. The Faster R-CNN algorithm was used for vehicle detection and the DeepSORT algorithm was used for tracking to obtain the vehicle trajectory. DeepSORT is an object tracking framework and is an extension of SORT (Simple Real time Tracker) [Hou *et al.*, 2019]. With the detection and tracking results, the MoG2 model was presented, based on a Gaussian mixture model [Reynolds, 2009], which aims to remove moving vehicles and only analyze stationary vehicles. Then, a new mask extraction mechanism is implemented based on the difference in frames and the vehicle's tracking trajectory to remove secondary roads, such as parking lots, and thus avoid false detections. Next, a multi-granularity tracking structure is used that contains a box-level tracking branch and a pixel-level tracking branch. Each branch contributes to capturing abnormal abstractions at different levels of granular-ity to model abnormal concepts. Finally, a backtracking and anomaly fusion optimization method is proposed to refine the abnormal predictions, which can significantly improve the robustness and accuracy of the results.

In [Zhao, 2021], a simple and efficient structure is proposed that includes three steps: Pre-processing, A dynamic track module, and Post-processing. The pre-processing step aims to generate candidate anomalies and is composed of four parts: video stabilization, background modeling, vehicle detection, and mask generation. Video stabilization aims to correct camera movement oscillations that occur in the acquisition process, through software techniques such as point matching based on Good Features to Track (GFTT) [Shi *et al.*, 1994] and a sparse optical flow is calculated to generate frame-by-frame transformations, and thus improve visual quality and improve end applications such as vehicle detection and tracking. In background modeling, a background subtraction approach based on Gaussian mixture (MoG) is used and the objective is to compare the difference between the results of forward and backward background subtraction, which shows that the results of backward background subtraction will make stationary vehicles clearer, and background-to-front subtraction is used as an auxiliary method to obtain a more accurate start time of the anomaly. Faster R-CNN is used to detect vehicles. When generating a mask, it is necessary to filter static vehicles on secondary roads and parking lots to detect anomalies on primary roads and for this, it will be necessary to segment the hypothetical anomalous regions of the mask using a mask extraction method based on movement and a mask based on trajectory. The dynamic tracking step searches for and locates the onset time of anomalies, using vehicle movement patterns and spatio-temporal status. Finally, post-processing is used to fine-tune the time limit of anomalies.

The related works described above serve as a direct influence on the creation of the proposal for this work. Table 1 presents the main comparisons with the works cited. Each one contributed with ideas that can be applied to vehicle detection, and vehicle tracking, where our proposed work focuses, as well as other articles that contribute to the detection of traffic anomalies.

In [Bafghi and Shoushtarian, 2020]'s work presented two distinct methods for extracting object features. The first method used SIFT and color histogram features in each image to evaluate the similarities between neighboring frames and different objects. On the other hand, the second method employed deep features obtained from the Mask R-CNN object detection network to achieve the same objective. In this work, we take bipartite graphs as a guide for object tracking but choose OPEN CLIP and structural similarity to determine connection weights. Unlike SIFT, which requires color histograms to enhance object features, OPEN CLIP can extract features effectively in various scenarios as it is designed to understand and correlate visual and linguistic information, enabling machines to process visual and text together. This ability to associate text and images makes it a valuable tool in several applications. Furthermore, structural similarity plays an important role when comparing and evaluating the quality of images resulting from different processes and transformations. It offers an objective metric to measure the structural similarity between original images and their modified versions.

In [Li *et al.*, 2020]'s work presented a solution to detect traffic anomalies using computer vision methods such as background subtraction, image segmentation, and modularized components to track vehicles at box and pixel levels. In this work, we use the idea of background subtraction and segmentation to remove vehicles on secondary roads. However, we chose to use Convex Hull to generate areas of both movement and lack of movement. We choose busy areas to analyze vehicle behavior and identify possible vehicles stopped on these roads. An important distinction between this work and the previous one is that, in our method, motion areas are generated during the tracking process. In the previous work, background subtraction and segmentation were performed separately in two different processes, which resulted in a higher computational cost.

In [Bai *et al.*, 2019], a solution for detecting traffic anomalies was presented, analyzing anomaly events based on vehicle dynamic information. Specifically, they used six spatiotemporal information matrices to identify the start and end time of the detected anomaly. This information was related to the pixel level and was associated with time. Our work incorporates time-related information to determine when a vehicle remains on the scene. To do this, we monitor whether the vehicle is still present at the scene or has left it. If the vehicle is in the scene, we calculate its speed to determine whether it is stopped or moving. When the speed is low, we interpret that the vehicle is stationary, and, in this case, we create hypothetical positions for this object, aiming not to lose important information about it, considering it as an object with a possible anomaly. A distinction between our method and the previous method is that our process incorporates temporal information using a unified approach.

In contrast, the previous method uses background modeling and perspective detection as separate processes to obtain information about the start and end of potential anomalies. Our method offers a more integrated approach, using information from vehicle trajectories to identify the beginning and end of possible anomalies. The results in a more complete and efficient vehicle behavior analysis in a single process.

Finally, in the work by [Zhao, 2021] presents three stages to detect traffic anomalies: pre-processing, dynamic tracking, and post-processing. In our work, we are inspired by the pre-processing stage, as we face challenges related to noise in the data, such as camera instability and variations in lighting. However, we chose to approach anomaly detection differently, using object tracking in conjunction with a data structure known as QuadTree, in addition to a temporal approach. The QuadTree is employed to compare the positions and characteristics of nearby objects, while the temporal structure analyzes whether these objects fall into the anomaly category. This approach makes our method robust in detecting anomalies compared to the other method, which requires additional post-processing to adjust the temporal boundaries of anomalies. In short, our strategy simplifies the detection process and improves the effectiveness of identifying traffic anomalies.

Therefore, the related works mentioned above are the leading guide for this project. They were developed from differ-ent perspectives, significantly enriching the knowledge necessary to define the method adopted in this research.

**Table 1.** Comparison between related works and the proposal of this paper

| Proposals | Dataset | Detection | Tracking | Traffic anomalies |
|---|---|---|---|---|
| [Bafghi and Shoushtarian, 2020] | UA-DETRAC | *Mask* R-CNN | Appearance model and Visual Object Tracking | — |
| [Li *et al.*, 2020] | *Track*4 | *Faster* R-CNN | — | Box and Pixel Level Tracking Model |
| [Bai *et al.*, 2019] | *Track*3 | *Faster* R-CNN | — | Perspective Relationship Detection Model |
| [Zhao, 2021] | *Track*4 | *Faster* R-CNN | — | Dynamic Tracking Model |
| This paper | Different datasets | YOLOv7 | Vehicle Tracking Using areas of interest | Area Anomaly Detection Space-Temporal Interest |

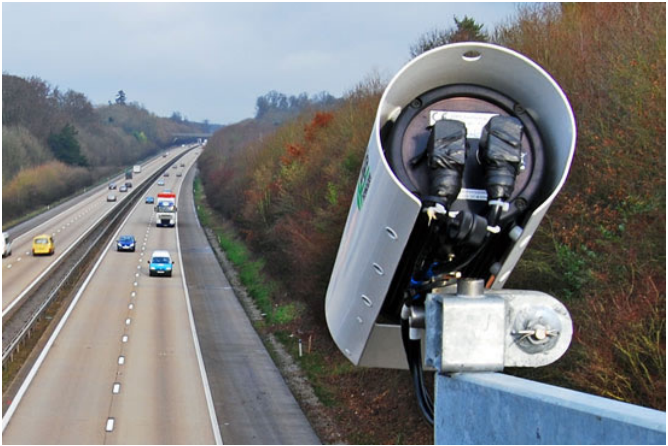

**Figure 1.** MEDAVET application scenario.

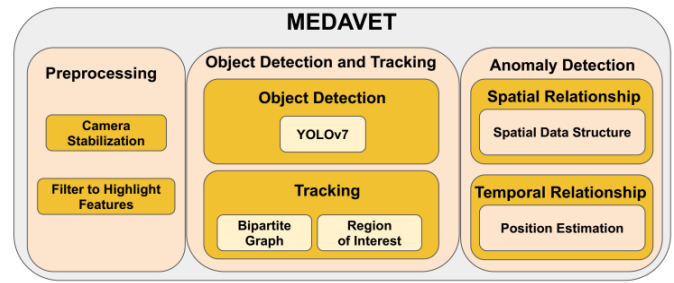

**Figure 2.** MEDAVET Overview.

# 3 MEDAVET: Traffic Vehicle Anomaly Detection Mechanism based on spatial and temporal structures in vehicle traffic

This section presents an anomaly detection mechanism called MEDAVET - Traffic Vehicle Anomaly Detection Mechanism. MEDAVET is based on detecting and tracking vehicles on urban roads, aiming to detect anomalies in highway traffic. Our approach focuses on identifying vehicles that remain stationary on main road lanes for more than one minute. In the road context, a "main road" refers to a road of greater capacity and importance within a road network. It is generally designed to accommodate a large volume of traffic. Our research aims to significantly improve road safety and traffic efficiency by identifying anomalous situations that may pose risks, such as prolonged congestion or incidents that require rapid intervention.

Therefore, in this paper, we consider that cameras monitor the highways along the highways (Figure 1) and that the information captured is sent to a center where processing is carried out. Thus, MEDAVET runs in the control center, allowing an operator to take action on time, such as calling a vehicle to check what is happening to the driver.

The following sections describe the components and func-tionalities that makeup MEDAVET.

## 3.1 MEDAVET Overview

In this study, we implemented a video vehicle monitoring system comprising a video preprocessing process for camera stabilization and enhancement of visual characteristics. We then use the preprocessed frames to detect and track vehicles (Figure 2). To do this, we use an object detection tool to detect vehicles in the captured frames. Therefore, an object in this work would be a vehicle in the image. After object detection, we use graph theory to create a structure for representing and analyzing vehicle trajectories, allowing vehicles to be tracked over time. Based on the output graph in the vehicle detection and tracking component, we perform anomaly detection that will check the time that the vehicle will be out of mobility and the place that the vehicle is stopped to perform the inference, whether it is normal behavior or not. In the following subsections, we will describe each component in more detail.

## 3.2 Preprocessing

The image capture process can result in unwanted noise, which manifests in undetected frames, unwanted camera movement, and the appearance of artifacts, among other problems, as illustrated in Figure 3. These noises can impair the proper functioning of vehicles, causing interference in the detection and tracking process. It can lead to multiple detections and IDs assigned to vehicles, compromising the accuracy and consistency of results. Therefore, it is essential to carry out a pre-processing process to mitigate these issues. When analyzing our data, we identified the need to apply camera stabilization and noise filtering techniques to highlight relevant features in images.

**Figure 3.** Imagens sem preprocessamento.

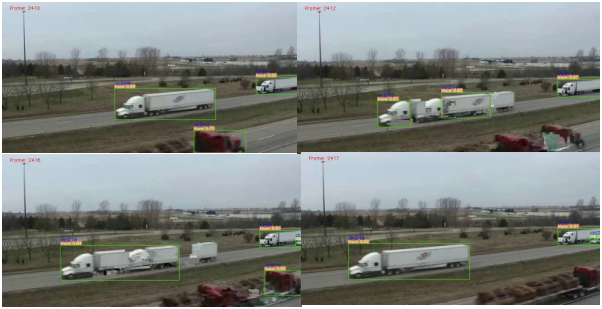

Camera stabilization is an essential process for improving the visual quality of videos, reducing the unwanted effects of camera movement. In this method, the goodFeaturesTo-Track algorithm detects and selects characteristic highlights in a frame, which will serve as references- key for subsequent tracking 4). The Lucas-Kanade Optical Flow algorithm, in turn, is used to track these characteristic points in subsequent frames, allowing precise determination of the displacement of these points over time. Based on these displacements, it is possible to calculate the rigid (Euclidean) transformation, which encompasses translation, rotation, and scale information, to correct the camera movement in each frame, resulting in a smoother and more stable final video.

**Figure 4.** Camera stabilization block diagram.

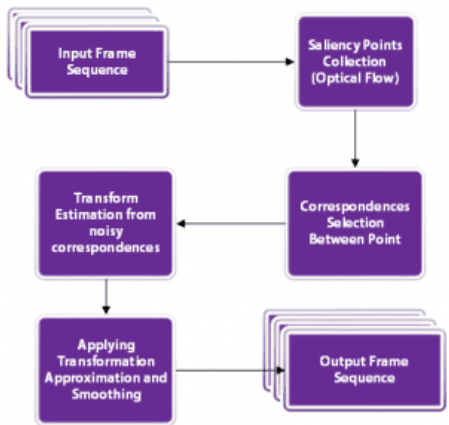

We take advantage of the inter-frame motion estimation performed in the previous step to filter noise in the motion trajectory. In this step, we seek to determine the movement trajectory by incrementally accumulating the estimated differential movement between consecutive frames. This means adding up the movement between frames to calculate the overall trajectory. The ultimate goal is to smooth this trajectory to make it more stable. For this smoothing, we use a moving average filter, which, as the name suggests, replaces the value of a function at a point with the average of the values of its neighbors. We apply this smoothed trajectory to obtain smoother motion transformations that can be applied to video frames to stabilize them. This is achieved by finding the difference between the smoothed and original trajectories and then adding this difference to the original transformations. The process involves iterating through the frames and applying these transformations, resulting in a final video that is stabilized and free from unwanted movements.

### 3.3 Vehicle Detection and Tracking

The object detection and tracking system is divided into two fundamental stages. We start by detecting objects in each frame of the videos using advanced computer vision algorithms, such as convolutional neural networks. We then create bipartite graphs to connect corresponding detections between consecutive frames, allowing continuous tracking of moving objects. To further improve efficiency, we delimited an area of interest around moving vehicles. Finally, the system assigns IDs to the tracked vehicles and records their trajectories over time. These trajectories provide valuable information for analyzing and understanding vehicle behavior in the context of the video (Figure 5 ).

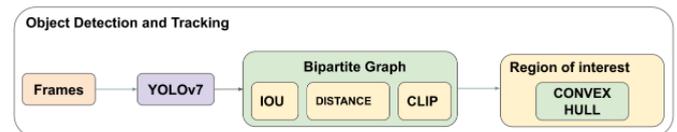

**Figure 5.** Detection and Tracking Component.

Initially, the objects in the image are detected; we use the YOLOv7 algorithm to perform object detection. The choice of YOLO v7 is due to its implementation of a new training algorithm called CrossEntropyLossWithLogits, which stands out for being faster and more efficient than the algorithms used in previous versions of YOLO. This optimization results in significantly reduced training time. Furthermore, this version incorporates different weights, trained on a vast image dataset, and can detect various objects, including vehicles. For this project, we used the YOLOv7-W6 model, which has proven highly effective in our quest for accurate object detection across various sizes, from minor to large-scale objects.

After object detection, the bipartite graph modulation begins, allowing the objects to be efficiently tracked over time and establishing connections between detections in different frames. For this, we consider that:

- Each vertex in one of these graphs represents a specific detection of an object in a frame. Therefore, if we have frames $F1$ and $F2$, each vertex in $F1$ represents an object detection from that frame, and each vertex in $F2$ represents a corresponding detection in the subsequent frame $F2$.
- Each set of vertices represents a video frame; that is, each vertex represents a vehicle detected by the YOLO algorithm.
- The weights are used to determine how the edges will be connected between the vertices of each consecutive frame and the bipartite graph.

Therefore, for every two consecutive frames, we create a bipartite graph so that all components in the current frame are connected to components in the next frame. Thus, the set of vertices $F$ is divided into $n$ disjoint sets, each representing a frame, and the components in each set represent the vehicles (detected objects).

For a better understanding, let us explain it in mathematical terms:

$$F = \{F_i \mid i \in \{1, \ldots, n\}, F_i \text{ disjuncts}\} \quad (1)$$

where $F_i$ represents the $i$-th frame of the set $F$. Then:

$$F_i = \{f_c^i \mid c \in \{1, \ldots, p\}\} \quad (2)$$

is the set of vertices of frame i, where $f_c^i$ denotes the $c$-th vertex. Having the set of frames and the set of vertices $p$, the set of edges can be defined as:

$$G = \{(f_c^i, f_q^{i+1}) \mid c \neq q\} \quad (3)$$

where $f_c^i$ represents the $c$-th vertex of frame i and $f_q^{i+1}$ represents the $q$-th vertex of frame $i + 1$

To calculate the weights, we use *IOU* (Intersection Over Union) metrics and feature extraction, with the aim of ensuring that the edges only connect vehicles with high similarity.

This is how the metrics are defined:

$$IOU(f_c^i, f_q^{i+1}) = \frac{Area((f_c^i) \cap Area(f_q^{i+1})}{Area((f_c^i) \cup Area(f_q^{i+1})} \geq 0.4 \quad (4)$$

To increase precision and assist the OR method, we also use the distance between the components of each vertex, giving the equation below:

$$dist(f_c^i, f_q^{i+1}) = \sqrt{(x_2 - x_1)^2 + (y_2 - y_1)^2} \leq 5 \quad (5)$$

where $(x_1, y_1)$ is the position of the center of $f_c^i$ and $(x_2, y_2)$ is the position of the center of $f_q^{i+1}$

To determine the similarity between the vehicles, we chose to use OpenAI CLIP, as it belongs to the Transformers family of models. Thus, the similarity metric we employ is defined as follows:

$$sim(f_c^i, f_q^{i+1}) \geq 0.7 \quad (6)$$

Combining the three parameters we have:

$$Weight(f_c^i, f_q^{i+1}) = \alpha * IOU(f_c^i, f_q^{i+1}) + \beta * sim(f_c^i, f_q^{i+1}) \quad (7)$$

With $\alpha$ and $\beta$ measurement parameters, the weights are calibrated between 0 and 1.

Within this work, we are mainly focused on tracking areas of interest along roads where road accidents occur, some of which may be caused by vehicle movement. Our main area of interest covers these specific areas of road movement. The goal is to continuously monitoring traffic in these regions to capture events such as accidents, incidents, or driving behaviors that could lead to dangerous situations. We use the Convex Hull algorithm to demarcate this area of interest in the image. This algorithm creates convex polygons surrounding the moving areas of the highway lanes. The Convex Hull is a convex envelope encompassing a set of points in the plane or a multidimensional space.

Creating these polygons allows us to clearly define the regions where moving vehicles are present, as illustrated in Figure 6. This way, we can exclude areas beyond the shoulder, such as gas stations and other establishments adjacent to the highway, focusing our analysis on areas directly related to traffic flow. This will help avoid confusion with vehicles on secondary roads. Secondary roads on a highway normally have less traffic capacity and are intended for specific purposes, such as access to parking lots.

**Figure 6.** Vehicles within the region defined via ConvexHull.

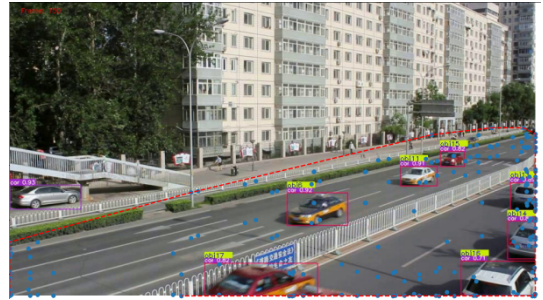

### 3.3.1 Functioning of the detection and tracking method

It is essential to highlight that video frames undergo a pre-processing step before being passed to the YOLO algorithm. This pre-processing improves the quality of the images and the location of objects. After this preparation, the frames are forwarded to the YOLO algorithm responsible for object detection. The tracking phase begins with identifying objects in each frame, as shown in the 1. For each pair of consecutive frames, a bipartite graph is created. The vertices of each frame are analyzed to check whether they meet the conditions defined by the bipartite graph. The vertices that meet these conditions receive a unique ID (line 6 of Algorithm 1). Otherwise, the similarity, IOU, and distance conditions are checked (lines 8 to 10). If these conditions are met, the frame and vertex information are assigned to the corresponding object in the list (line 12). Otherwise, a new object is created (line 14). The values used in line 11 were objectified through experiments described in section 4.

To minimize the number of vehicles to be analyzed, we use the Convex Hull algorithm to surround the area of moving vehicles and create a new list of objects that are in the area of interest (list_IA_obj). Thus, if there is a vehicle in the area of interest (Lines 17 to 19), an ID is assigned to that vehicle, excluding vehicles or other objects that are outside that area.

### 3.4 Anomaly detection

The anomaly detection component aims to detect vehicles stopped on the main roads. We implemented two essential structures: one focused on spatial analysis and the other on temporal analysis (Figure 7). Spatial analysis performs spatial searches, allowing the locations of objects to be obtained, especially to identify stationary vehicles. On the other hand, temporal analysis is crucial in dealing with challenges like object occlusion, ensuring we do not lose information about the object. Temporal analysis makes it possible to identify

---

**Algorithm 1** Vehicle tracking

---

**Input:** frames
**Output:** list_IA_obj, trajectory_obj
1: i = 0
2: **while** $i \leq n\_frames - 1$ **do**
3:     Graf[i] = Get_obj($F_i, F_{i+1}$)
4:     obj = Graf[i].get_obj
5:     **if** Is_vértice($obj_i, obj_{i+1}$) **then**
6:         update(list_obj($obj_i, obj_{i+1}$)
7:     **else**
8:         IOU = IOU($obj_i, obj_{i+1}$)
9:         dist = dist($obj_i, obj_{i+1}$)
10:        sim = sim($obj_i, obj_{i+1}$)
11:        **if** $IOU \geqslant 0.4$ and $dist \leqslant 30$ and $sim \geqslant 0.6$ **then**
12:            update(List_obj($obj_i, obj_{i+1}$))
13:        **else**
14:            list_obj.add(obj)
15:        **end if**
16:    **end if**
17:    **if** Is_obj(i) and Is_AI(obj) **then**
18:        list_IA_obj.add(obj)
19:    **end if**
20: **end while**

---

objects that have left the scene or are immobilized on the road for a prolonged period or even when the detector does not detect them. This combined approach allows us to detect stopped vehicles on busy roads, providing a robust solution for identifying anomalies in road traffic.

**Figure 7.** Anomaly Detection Component.

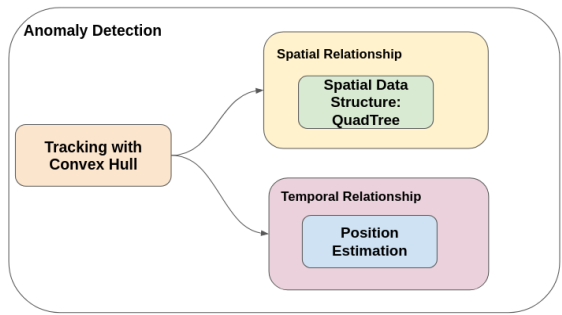

In this last step, data from objects (vehicles) within an area of interest calculated by Convex Hull is received. However, it was observed that when the vehicle is stopped at the scene, another vehicle can overlap the stopped vehicle, thus making a more accurate analysis difficult. Therefore, for the spatial component, we used the QuadTree method to improve spatial data organization within the delimited area. QuadTree performs a recursive division of the bounded region into smaller quadrants (Figure 8), creating a hierarchical representation of spatial information. This approach allows the analysis of vehicles that are close in terms of location and excludes regions where analysis is not necessary.

Furthermore, QuadTree plays an essential role in analyzing vehicle behavior. By hierarchically grouping spatial information, we can identify movement patterns, average speeds, vehicle interactions, and other essential aspects of vehicle behavior. This contributes to a deeper and more refined understanding of the traffic landscape, enabling effective anomaly detection.

**Figure 8.** QuadTree construction process

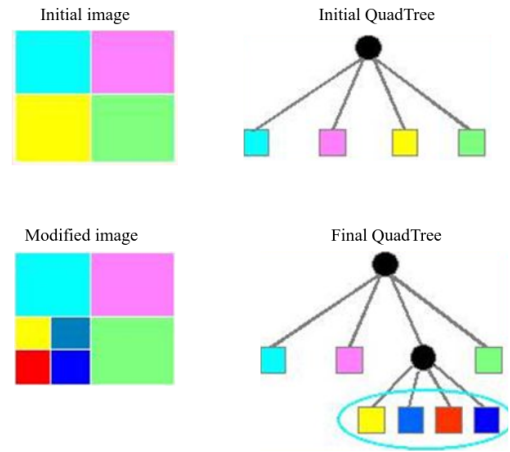

So, for the space stage, QuadTree proves a reduction in computational cost and in analyzing the behavior of vehicle trajectories, especially for stationary vehicles. It analyzes objects in the current frame about nearby objects in previous frames, considering a radius 'r' as a limit for comparing objects in previous frames. The QuadTree is divided into four subtrees, each representing a spatial region of the field of view, and chooses the region where vehicles are close to the vehicles being analyzed. This data structure improves vehicle tracking and is used with another data structure dedicated to temporal analysis.

On the other hand, the temporal data structure handles situations where vehicles may be temporarily out of the scene. It approximates the positions of temporarily unavailable vehicles, ensuring we keep information about them. Furthermore, it analyzes the time a vehicle is in the scene, which is essential to avoid reassigning existing IDs to new vehicles that enter the field of view. This approach prevents vehicles parked in the main scene from being mistakenly re-identified, allowing us to keep a complete record of these vehicles and analyze whether or not there are anomalies based on dwell time. This temporal analysis is essential to identify vehicles representing an anomalous situation, such as prolonged congestion or an unscheduled stop on a busy road.

To deal with the temporal mechanism, we use the vehicle speed in each frame, checking whether the vehicle is moving or stationary. If the vehicle is stopped, we start monitoring this object; if it remains stopped for around 1800 frames, corresponding to 1 minute that the vehicle will be stopped on the highway, it will be classified as a vehicle with abnormal behavior. A vehicle stopped on the highway can cause a severe accident.

### 3.4.1 Functioning of the anomaly detection method

The anomaly dating method receives the tracking of objects within an area of interest and their trajectory as input. With

---

**Algorithm 2** Anomaly Detection

**Input:** list_IA_obj, trajectory_obj
**Output:** list_ano
1: **for** veh in trajectory_obj **do**
2:     **if** veh in list_IA_obj **then**
3:         quadtree = QuadTree(veh)
4:         IOU = IOU(quadtree.obj)
5:         dist = dist(quadtree.obj)
6:         sim = sim(quadtree.obj)
7:         **if** $IOU \geqslant 0.4$ and $dist \leqslant 30$ and $sim \geqslant 0.6$ **then**
8:             update(List_Ia_obj(quadtree.obj))
9:         **else**
10:           **if** Is_scena(quadree.obj) **then**
11:              update(List_Ia_obj(quadtree.obj))
12:           **else**
13:              **if** Velocity_Zero(quadtree.obj) **then**
14:                 **if** Is_TheSameID(quadtree.obj) **then**
15:                     timestamp ++
16:                 **end if**
17:              **end if**
18:           **end if**
19:         **end if**
20:         **if** timestamp ==1800 **then**
21:             list_ano.add(quadtree.obj)
22:         **end if**
23:     **end if**
24: **end for**

---

this information, the anomaly detection method will use the QuadTree to analyze nearby objects in previous frames (line 3 of Algorithm 2). To analyze proximity, an analysis of similarities and distance threshold criteria and IOU is carried out with previous objects in previous frames about the current frame (lines 4 to 7). The object information is updated if these criteria are met (line 8).

However, if the criteria are not met, the timing structure checks the object's position to determine whether it is still in the scene. To do this, we construct a rectangle and check if the object is inside. If he is not, we assume he has left the scene (lines 9 to 12). However, if it is present, we analyze its speed over the last 100 frames. If this speed is deficient, almost zero, we infer that the vehicle is stopped at the scene. In this case, we understand that the vehicle is hidden, and in order not to lose information about its trajectory, we make hypothetical estimates in the next frame (lines 13 to 14).

Again, we use QuadTree to evaluate the proximity and similarity of the approximated object to nearby objects in previous frames, keeping its ID along with all previous information. Finally, we consider the number of frames in which the vehicle was in the scene. If this quantity exceeds 1800 frames, we classify it as an anomaly (lines 18 to 19).

To make the algorithm more understandable, imagine a set of consecutive frames. In the first frame, we can visualize a stopped vehicle and another vehicle overlapping the stopped vehicle (Figure 9). The analysis process starts with the first frame, using the QuadTree to examine the nearby objects in the previous frames. QuadTree is a technique that recursively divides the image region into smaller quadrants, creating a hierarchical representation of the spatial informa-tion of objects. We use similarity, distance, and IOU metrics to evaluate whether current objects are similar to objects in previous frames. If the current objects do not resemble the nearby objects in the previous frames, we start using the temporal data structure to check if the vehicles are still in the scene.

We define a rectangle inside the frame, with margins of -10 pixels on each side of the rectangle. If vehicles are within this rectangle, we assume they are still in the scene (frames two and three) but may not have been tracked correctly. In this case, we estimate their trajectories by calculating their speeds over the last 100 frames. If the speed is deficient, indicating that the vehicle is practically stopped, we assume it is hidden in the scene.

After estimation, we apply QuadTree again, using similarity, distance, and IOU metrics, to evaluate the proximity and similarity of the estimated object with nearby objects in previous frames. We keep the same ID and all previous object information throughout this process (last frame). To classify the situation as an anomaly, we count the number of frames in which the vehicle was present in the scene. If this quantity is more significant than 1800 frames, we consider the situation an anomaly. This approach allows us to identify situations where a vehicle remains on scene for an unusually long time, which may indicate an anomaly.

**Figure 9.** Ilustração da relação temporal e espacial.

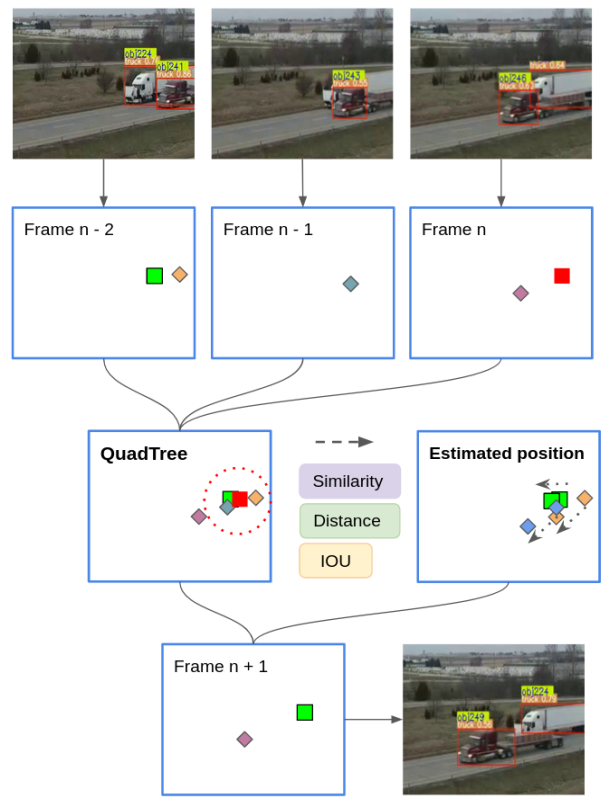

## 4 Analysis of Results

In this section, we will describe in detail the conduct of the experiments, presenting specific information about the proposed method for detecting, tracking, and detecting anoma-lies in vehicles. This method incorporates an innovative approach that combines computer vision, machine learning, and deep learning algorithms to detect vehicles, track their trajectory, and identify anomalous behaviors, thus contributing to a safer and more efficient environment in the context of transport and mobility.

## 4.1 Implementation Details

This work was implemented in the Python programming language, using version 3.11, and executed on a machine with the Linux operating system (Ubuntu 20.04). The hardware used to support development includes an 8-core Intel Core i7 processor, 16GB of RAM, and an NVidia RTX-3060 card with 12GB of video memory. This hardware configuration provided efficient performance during all phases of work implementation, guaranteeing the necessary processing capacity for the tasks of this work.

## 4.2 Object Tracking

The UA-DETRAC [Wen *et al.*, 2015] dataset was used to implement the tracking method. UA-DETRAC is a challenging benchmark for evaluating vehicle detection and tracking algorithms. Composed of 10 hours of videos recorded on the streets of China, this dataset includes a total of 140,000 frames, recorded at a rate of 25 frames per second (fps) and with a resolution of $960 \times 540$ pixels. This dataset covers the presence of 8,250 vehicles distributed across all frames.

### 4.2.1 Assessment Metrics

Evaluation metrics are intended to measure performance in the analysis of object-tracking systems. This work focuses on two essential areas: multi-object tracking and vehicle speed estimation. We use the CLEAR MOT metrics to evaluate object tracking, as proposed by [Bernardin and Stiefelhagen, 2008], which has several performance aspects. The principal metric we employ is MOTA (Multi-Object Tracking Accuracy), which is fundamental for the overall assessment of multi-object tracking performance. MOTA takes into account the number of false positives (FP), false negatives (FN), and identity changes (IDS) from Ground Truth (GT). In essence, MOTA quantifies how well our tracking system correctly detects and follows objects compared to actual annotations. A higher MOTA value indicates superior tracking performance with fewer object detection and identification errors. The formula for calculating MOTA is as follows:

$$MOTA = 1 - \frac{\sum_t (FN_t + FP_t + IDS_t)}{\sum_t GT_t} \quad (8)$$

Furthermore, we use metrics such as MOTP (Multi-Object Tracking Precision), which characterizes the misalignment between the annotated and predicted bounding boxes. MT (Mostly Tracked) and ML (Mostly Lost) are used to evaluate the continuity of the trajectory of tracked objects. Furthermore, identity switches (IDS) are used to measure the tracking quality regarding object identification. To evaluate computational performance, we also consider the number of frames processed per second (Hz), which describes the overall inference speed of the system.

### 4.2.2 Parameter Evaluation

To determine the appropriate values for the parameters of our method, we carried out a series of experiments, exploring different values of similarity, distance, and IOU, as seen in Table 2. We start with fixed settings, keeping a high and constant value for similarity (Sim) and a low and continuous value for distance (Dist), while varying the IOU values in the first three rows. However, in the last three lines, we decided not to fix specific values for these parameters, allowing us to observe the considerable impact of distance, significantly when the IOU is reduced due to the significant size of some vehicles and their high speed. The most promising results from our experiments are highlighted in the last row of the table. Notably, it is observed that false negatives (FN) and identification misses (IDS) are lower compared to the other configurations.

However, it is essential to note that the false positives (FP) are slightly higher than in the penultimate line, with a minimum difference of just 4, making MOTA have a better result than the others. The goal is to minimize FP, FN, and IDS while maximizing the MOTA score. For our tests, we selected video 2 from the test set, consisting of 1120 frames. We used structural similarity as our feature extractor, mainly due to its shorter execution time. The weights of the bipartite graph were generated according to Equation 7, where $\alpha$ and $\beta$ are parameters that adjust the weights to ensure that they remain in the range between zero and one. In this work, we determined that the appropriate values for $\alpha$ and $\beta$ are, respectively, 0.4 and 0.6.

**Table 2.** Assessment of tracking parameters.

| Parameters | | | Results | | | |
|---|---|---|---|---|---|---|
| **Sim** | **IOU** | **Dist** | **FP ↓** | **FN ↓** | **IDS ↓** | **MOTA ↑** |
| 0.9 | 0.9 | 5 | 35 | 2346 | 27 | 0.7 |
| 0.9 | 0.8 | 5 | 276 | 485 | 200 | 60.4 |
| 0.9 | 0.7 | 5 | 390 | 122 | 11 | 78.4 |
| 0.8 | 0.6 | 10 | 395 | 126 | 7 | 78.2 |
| 0.7 | 0.5 | 20 | 378 | 129 | 5 | 78.9 |
| 0.6 | 0.4 | 30 | 382 | 125 | 4 | 78.9 |

### 4.2.3 General feature extractor results

The experiments on the UA-DETRAC test suite, using the OPEN CLIP, Structural Similarity (SSMI), and SIFT feature extractors, clearly demonstrated the superiority of OPEN CLIP over SSMI and SIFT. This superiority is reflected in significant improvements in metrics, including the reduction of false positives (FP), false negatives (FN), and erroneous identifications (IDS), as well as superior performance in terms of tracking accuracy (MOTP) and overall tracking metrics (MOTA). The results of these experiments are detailed in Figure 10.

This advantage of OPEN CLIP is attributed to its unique ability to analyze the semantic structure of each object, making it notably more robust than other extractors. Specifically, compared to MOTA, OPEN CLIP outperformed SSMI by 11.5% and SIFT by 33.7%, demonstrating its effectiveness in the object tracking task. It is important to note that, concerning MOTP, the OPEN CLIP presents a comparable performance to the other two extractors. This occurs because MOTP evaluates the spatial difference between the bounding boxes of the real data and those generated by the proposed method, and the Yolo algorithm plays a fundamental role in standardizing and providing this information, allowing a fair comparison between extractors. It is essential to mention that, in terms of processing rate (Hz), SSMI and SIFT surpassed OPEN CLIP. This is because OPEN CLIP is more computationally intensive due to its detailed semantic analysis. However, considering the importance of tracking accuracy and performance, choosing OPEN CLIP as a feature extractor is justified.

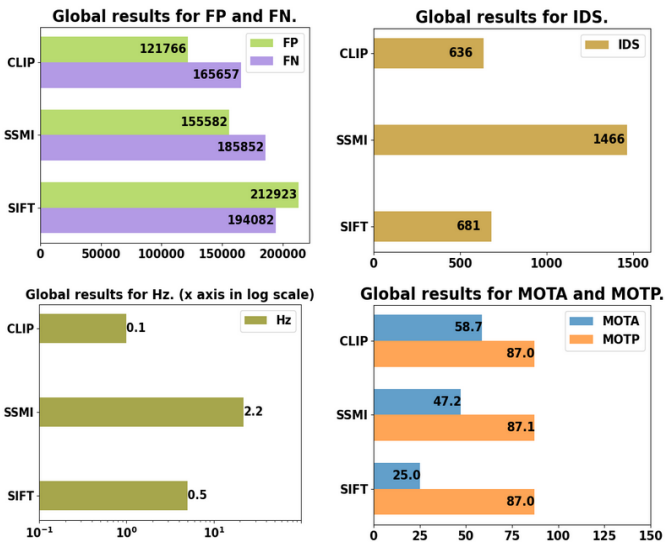

**Figure 10.** Feature extractor results.

#### 4.2.4 General results of MEDAVET with other methods

Assessment of MEDAVET's performance compared to other models in the literature is important to determine its effectiveness and relevance. I will provide a brief description of seven models found in the literature to make a comparison with them.

SORT [Bewley *et al.*, 2016] is an algorithm in real-time object tracking. It is based on using particle diffusion, a probabilistic approach, to estimate position and track objects in video sequences. One of the advantages of SORT is its simplicity, which makes it efficient for real-time use. It is beneficial in scenarios where it is necessary to track moving objects on video, such as in surveillance systems, autonomous vehicles, and video analytics. SORT may be less accurate in challenging situations, such as when objects are very close to each other or when occlusions occur.

IOU [Bochinski *et al.*, 2017] is an object tracking algorithm notable for its ability to track high-speed objects with high accuracy, all without the need to process image information. Instead, it relies on a simple motion model, making assumptions about objects' movements. The IOU's success is due to its efficiency and ability to handle high-speed objects. As it does not rely on video image analysis, it is computationally efficient and suitable for scenarios where processing speed is critical. However, it is worth noting that the IOU may be less robust in scenarios where the assumptions of straight-line motion and constant speed do not apply. It may also be less effective in scenarios with frequent occlusions or when objects abruptly change direction.

CMOT [Bae and Yoon, 2018] is an online object tracking algorithm that stands out for its robustness and effectiveness, especially in complex scenes with multiple objects and objects of similar appearance. It uses deep learning techniques to discriminate objects that share close visual characteristics and divides the object tracking problem into smaller subproblems based on the confidence of the tracks; this allows the algorithm to track objects accurately and efficiently, even in complex scenes with many objects.

Model2 [Munjal *et al.*, 2020] is a joint object detection and tracking algorithm in videos using identification features. The algorithm uses a deep learning model to detect and track objects simultaneously. The deep learning model is trained on a video dataset that contains objects labeled with identifying characteristics such as color, texture, and shape. The algorithm improves the performance of both tasks (detection and tracking) compared to conventional approaches. The algorithm uses identifying characteristics to associate object detections with object tracks accurately. The algorithm is also efficient and robust to occlusion, lighting variations and changes in the appearance of objects. This is because the deep learning model is trained on a challenging video dataset that contains a wide variety of objects and conditions.

JDE [Wang *et al.*, 2020] is a real-time multiple object tracking algorithm that stands out for its ability to perform both object detection and tracking simultaneously. It uses a deep neural network to learn these tasks jointly, resulting in accurate and efficient tracking, even in complex scenarios with multiple objects.

FairMOT [Zhang *et al.*, 2021b] is a multi-object tracking algorithm that treats detection and re-identification in a balanced way. This is important because detection performs better in conventional approaches than re-identification, as it is easier. Task imbalance can lead to lower overall performance of the tracking algorithm. The FairMOT algorithm uses a unique neural network to simultaneously perform detection and re-identification tasks. This avoids the task imbalance problem and allows the algorithm to achieve state-of-the-art results on challenging object-tracking datasets. The FairMOT algorithm is also efficient and robust to occlusion, lighting variations, and changes in object appearance. This is because the unique neural network is trained on a diverse and challenging video dataset.

ECCNet [Yu *et al.*, 2022] is an advanced multi-category vehicle real-time tracking and speed estimation algorithm.

ECCNet uses an efficient deep neural network composed of three main modules: detection, tracking, and speed estimation. Its chained structure allows efficient reuse of feature map features, reducing computational cost and improving performance in challenging situations such as occlusions and lighting variations. This system offers a robust and effective solution for applications that require accurate real-time vehicle tracking, representing a significant advancement in computer vision.

The detailed experimental comparison on the test set of the UA-DETRAC dataset, as summarized in Figure 11, demonstrates that our tracking approach significantly outperforms several existing methods. MEDAVET achieves a MOTA of 58.7%, considerably outperforming methods such as SORT, IOU, CMOT, Model2, JDE, FairMOT, and ECCNet by 42.3%, 39.3%, 46.1%, 3.6%, 34.2%, 27% and 3.2%, respectively. This is due to CLIP's excellent feature extraction performance and ability to work effectively in areas of interest. MEDAVET performs comparable to ECCNet and Model2 concerning MOTP. This is because these three models use advanced detection methods, resulting in detection bounding boxes that are close to real objects. CMOT and FairMOT outperform MEDAVET in terms of ID Switches.

These methods employ re-identification models that can associate object detections based on unique characteristics, differentiating similar objects even in complex scenarios. On the other hand, MEDAVET uses a detection-based association method, which may result in association errors in challenging situations. MEDAVET is only outperformed by Model2 and ECCNet over MT and by Model2 over ML. This indicates that MEDAVET is highly competitive in keeping track of most objects, but there is still room for improvement regarding lost objects. MEDAVET is surpassed in terms of processing rate by other models due to CLIP's computational complexity and detailed semantic structure analysis. It is essential to highlight that, in applications where precision is fundamental, MEDAVET is justified, even if it is less fast than other models.

## 4.3 Anomaly Detection

To verify the performance of the anomaly detection method, the track4 dataset from NVIDIA AI CITY CHALLENGE 2021 was used. The track4 is divided into a training set and a test set. The training set has 100 videos with a duration of approximately 15 min, and the test set has 150 videos the same length as the training videos. The videos in the dataset have a resolution of $1920 \times 1080$ *pixels*, recorded at a rate of 30 *frames* per second (fps).

### 4.3.1 Assessment metrics

To evaluate the anomaly detection performance on the test set, the $S4$ metric is used, which is the combination of two metrics: the F1 score and the normalized root mean square error (NRMSE).

$$S4 = F1 \times (1 - \text{NRMSE}) \qquad (9)$$

The F1 score is the harmonic mean of recall and precision. Specifically, a true positive (TP) detection is considered the correct anomaly within ten seconds of an actual anomaly (before or after). A false negative (FN) is a real anomaly that our algorithm cannot correctly predict. A false positive (FP) represents the predicted anomaly but is not a real anomaly. The F1 score is summarized by:

$$F1 = \frac{2\text{TP}}{2\text{TP} + \text{FN} + \text{FP}} \qquad (10)$$

**Figure 11.** Resultados gerais com outros métodos.

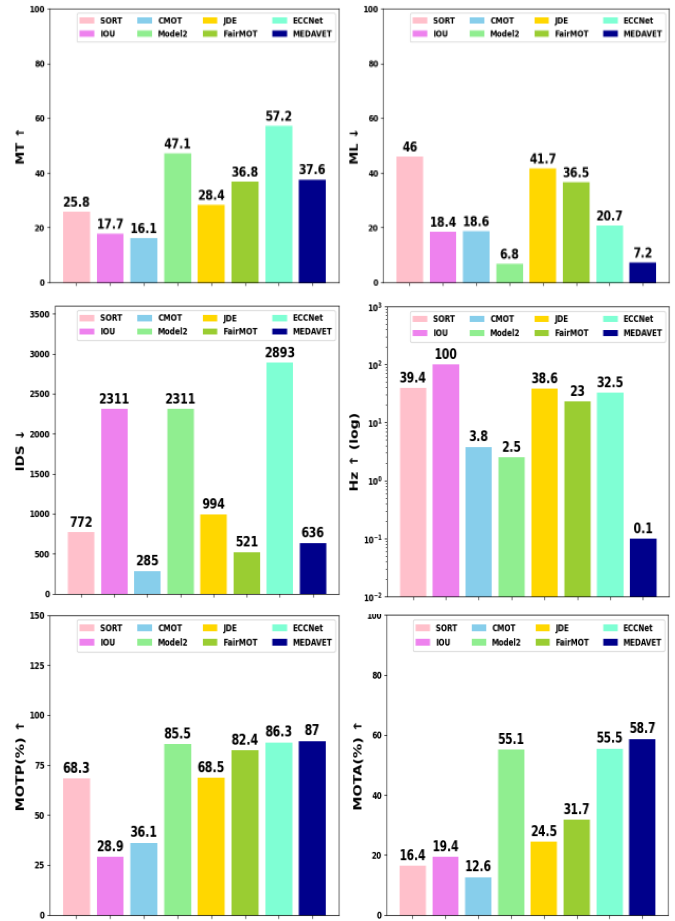

The normalized root mean square error (NRMSE) denotes the temporal error of the predicted time (by our method) and the ground truth time for all true positive predictions. NRMSE employs a max-min normalization with a maximum value of 300 and a minimum value of 0. In short, NRMSE is defined as follows:

$$\text{NRMSE} = \frac{\min\left(\sqrt{\frac{1}{\text{TP}} \sum_{i=1}^{\text{TP}} \left(t_i^p - t_i^{gt}\right)^2}, 300\right)}{300}, \qquad (11)$$

where $t_i^{gt}$ denotes the start time of the ground truth anomaly and $t_i^p$ is the predicted start time proposed by our method.

### 4.3.2 General results of MEDAVET in anomaly detection

The algorithm identifies the largest number of anomalies present in the 150 videos in the test set. It provides the start and end time of the anomaly, in addition to the confidence score.

Unidentified anomalies are attributed to challenging video scenarios, such as adverse weather conditions such as fog or nighttime periods where the detector's capabilities are limited. Furthermore, the distance between the camera and the vehicles, together with the small size of the vehicles, contribute to the constant non-detection of these vehicles. Figure 12 We present a visual illustration of these scenarios. In the first three frames of video 45 of the test set, the detector cannot always identify the vehicles, resulting in the loss of information and, consequently, the non-detection of the anomaly. In the following three frames of video 41, also from the test set, the nighttime scenario makes detecting vehicles that are stopped for long periods even more challenging, contributing to the non-detection of anomalies.

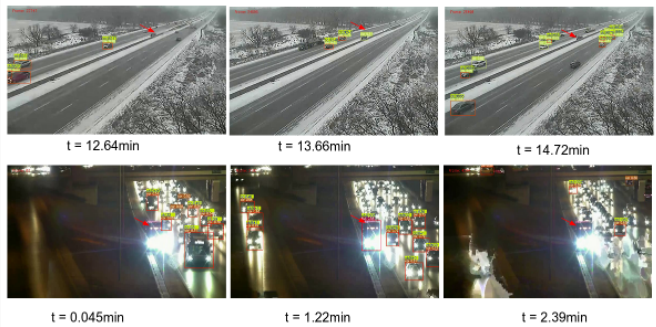

**Figure 12.** Undetected anomalies.

Figure13 presents frames from video 43 of the test set, allowing the visualization and illustration of the anomalies discussed in our study. To distinguish vehicle stopping periods, we adopted a color signaling system. An alarm signal is displayed in green, indicating that the vehicle has been stopped for one minute. After this period, the alarm changes to yellow and remains in that color for two minutes. Finally, it turns red to indicate that the vehicle has been stopped for more than three minutes. The change of colors over time is intended to inform that the longer the stop, the greater the risk of accidents.

**Figure 13.** Anomalies detected

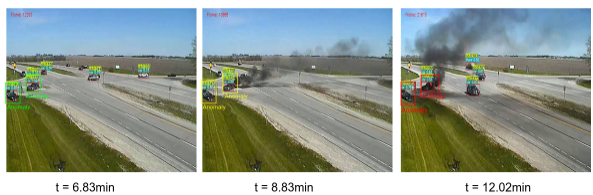

In the context of this work, the analysis of true positives, false negatives, and false positives is very relevant in evaluating system performance. Figure 14 illustrates the results of the entire test suite against these metrics. A true positive in this scenario indicates agreement between actual and predicted data, whether for anomalies or non-anomalies. In this context, we found a total of 198 true positives. On the other hand, false negatives indicate that, although the anomalies are present in the real data, the predicted data does not correctly identify them. Otherwise, the vehicle that presents anomalies in the real data does not coincide with the predictions, resulting in 13 false negatives. As far as false positives are concerned, they occur when predictions indicate anomalies that are not present in the actual data, and in this case, we found 12 false positives.

The presence of false negatives and false positives is undesirable in any scenario. However, if it were necessary to choose between the two, true positives are the least harmful option. This is because a true positive would lead to an anomaly alarm, which may result in loss of time for accident assistance personnel when traveling to the scene. However, there would be no real risk situation. However, the worst case scenario would be a false negative, as in this case, there would be a real anomaly, but the algorithm would not detect it, which could lead to severe accidents, including the risk of loss of life if appropriate assistance is not provided.

**Figure 14.** Overall results of TP, FN and FP.

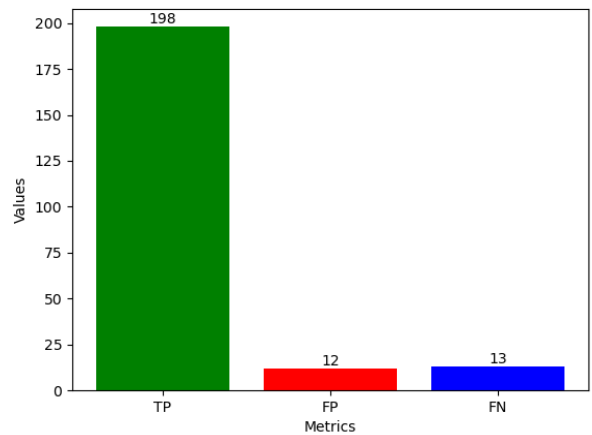

We evaluated the performance of our method on the NVIDIA AI CITY CHALLENGE 2021 Track4 test set. As evidenced in Table 3, we obtained an overall score of 0.7845 on the $S4$ metric, accompanied by a solid F1 score of 85.71%. Furthermore, the error in the start time was 25.432 seconds, which highlights the robustness of our proposed method.

**Table 3.** Anomaly Detection method result

| S4 | F1 | RMSE |
|--------|--------|--------|
| 0.7845 | 0.8571 | 25.432 |

# 5 Conclusion

This paper presents an innovative vehicle detection and tracking model, which uses bipartite graphs to model the tracking process. Furthermore, we incorporate the Convex Hull algorithm with the aim of clustering areas where vehicles are moving. To deal with vehicles that remain stationary for long periods and can be temporarily hidden, we employ the QuadTree data structure and a temporal structure, allowing us to group these vehicles and estimate their positions effectively. The results obtained are acceptable. Performance is evaluated based on a total score of 0.7845, F1 score of 85.71%, and NRMSE of 25.432. In future work, we can improve classifying objects and detecting anomalies in a darker scenario or with a boisterous image.

## Acknowledgements

A. R. H. Reyna would like to thank the Coordination for the Improvement of Higher Education Personnel (CAPES), Brazil for the financial support grant #PROEX-12636374/M and #88887.661865/2022-00. Alex J. F. Farfán acknowledges the financial support grant #PROEX-9527567/D from CAPES, Brazil. R. I. Meneguette thanks the support from CAPES, CNPq and the São Paulo Research Foundation (FAPESP) under grant no. #2020/07162-0 and #2022/00660-0.

## Funding

This work was supported by São Paulo Research Foundation (FAPESP) under grant no. #2020/07162-0 and #2022/00660-0, and collaboration of the CAPES.

## Authors' Contributions

All authors contributed to the writing of this article, read and approved the final manuscript

## Competing interests

The authors declare that they have no competing interests.

## Availability of data and materials

Data can be made available upon request.